\documentclass{article}
\usepackage{authblk}
\usepackage[utf8]{inputenc}
\usepackage{amsmath,amsfonts}
\usepackage{algorithmic}
\usepackage{algorithm}
\usepackage{array}
\usepackage[caption=false,font=normalsize,labelfont=sf,textfont=sf]{subfig}
\usepackage{textcomp}
\usepackage{stfloats}
\usepackage{url}
\usepackage{verbatim}
\usepackage{graphicx}
\usepackage{cite}
\usepackage[margin=25mm]{geometry}
\usepackage{amsmath}
\usepackage{amsfonts}
\usepackage{amssymb}
\usepackage{graphicx}
\usepackage{verbatim}

\begin{document}

\title{Stacked Cross-modal Feature Consolidation Attention Networks for Image Captioning}

\author{Mozhgan Pourkeshavarz, Shahabedin Nabavi, Mohsen Ebrahimi Moghaddam, Mehrnoosh Shamsfard}
\affil{Faculty of Computer Science and Engineering, Shahid Beheshti University, Tehran, Iran}
\affil{Corresponding Author:  Mohsen Ebrahimi Moghaddam, Email: m\_moghadam@sbu.ac.ir}


\providecommand{\keywords}[1]
{
  \small	
  \textbf{\textit{Keywords---}} #1
}


\maketitle

\begin{abstract}
Recently, the attention-enriched encoder-decoder framework has aroused great interest in image captioning due to its overwhelming progress. Many visual attention models directly leverage meaningful regions to generate image descriptions. However, seeking a direct transition from visual space to text is not enough to generate fine-grained captions. This paper exploits a feature-compounding approach to bring together high-level semantic concepts and visual information regarding the contextual environment fully end-to-end. Thus, we propose a stacked cross-modal feature consolidation (SCFC) attention network for image captioning in which we simultaneously consolidate cross-modal features through a novel compounding function in a multi-step reasoning fashion. Besides, we jointly employ spatial information and context-aware attributes (CAA) as the principal components in our proposed compounding function, where our CAA provides a concise context-sensitive semantic representation. To make better use of consolidated features potential, we further propose an SCFC-LSTM as the caption generator, which can leverage discriminative semantic information through the caption generation process. The experimental results indicate that our proposed SCFC can outperform various state-of-the-art image captioning benchmarks in terms of popular metrics on the MSCOCO and Flickr30K datasets. 

\end{abstract}

\section{Introduction}
Automatically describing the content of images, known as image captioning, is a significant task of artificial intelligence, which combines the field of computer vision (CV) with natural language processing (NLP). Image captioning has several applications for image indexing, social media platforms, visually impaired people, etc. Although this task seems easy for humans, it is complicated for machines. Machines should solve the problem of identifying which objects and attributes are present in the image, and their interactions must be expressed in natural language. The recent progress in deep neural networks has taken the latest significant step towards a reliable solution in generating descriptions for images.

In particular, deep image captioning architectures have shown impressive results in discovering the mapping between visual features and their correspondences in natural language. The well-known encoder-decoder framework is used to perform the task. It contains a convolutional neural network (CNN) for feature extraction and long short-term memory (LSTM) to generate a sentence based on the static overall image feature vector [1]-[5]. Although the advancements in these techniques are encouraging, a bottleneck facing the mentioned framework is that it is troublesome to mine all the visual information essential to construct a caption that accurately describes the image.

Inspired by the presence of attention in the human visual system that tends to focus on particular parts of the whole visual space [6], [7], visual attention has been proposed. Specifically, rather than encoding an image into a single static vector, visual attention encourages the model to selectively focus on salient areas of the image and use these areas to generate captions [8]. Besides, some models focus only on the salient regions without scanning the entire image, which cannot capture the context. Although that approach is interesting, it suffers from two main drawbacks, which motivate further significant research. The first is that the model generates visual words rather than high-level semantic words. The other problem is that they lack textual information, which leads to an inaccurate understanding of image context. Several studies have shown that the use of attribute-based methods aims to generate more advanced semantic details to boost the performance of image captioning [9]-[11]. However, the downside of this effective approach is that incorporating all the existing attributes in the image into the recurrent neural network (RNN) is unnecessary or even misleading. More recent evidence highlights that there is a need for both visual and semantic information due to their complementary nature [12], [13]. Despite the most significant benefit of their work, they do not consider the interrelationship between visual and semantic information and combine them to construct an abstract representation regardless of the image context.

Generally, generating a subjective sentence to describe the salient points in the image requires more abstract words. In many cases, inferring these words needs to consider more than one region in addition to high-level semantic concepts with an awareness of contextual information. For example, in the caption, "soccer fans cheer their team and celebrate the goal in a full stadium with open-air", it is surprising that none of the words can be classified only from a bounding box visual region. Intuitively, when predicting words like "fans" and "goal", the model must make inferences based on visual and semantic information under the umbrella of contextual representation. Therefore, using the visual regions to generate fine-grained captions is not enough. On the other side, involving high-level semantic concepts and contextual information along with the region of interest (ROI) needs an advanced feature fusion approach.

This paper sheds new light on cross-modal feature consolidation for the image captioning task. Specifically, we form our novel compounding function in the proposed stacked cross-modal feature consolidation (SCFC) attention network. In particular, with progressive reasoning via multiple CFC layers, the SCFC can gradually consolidate cross-modal knowledge to generate a rich representation through the caption generation process in every time step. We also construct our cross-modal features as principal components in the compounding function to boost the captioning result. Precisely, spatial-visual information, high-level semantic concepts, and contextual information are all considered for preparing an abstract and richer representation of the given image. To better use consolidated features potential, we further offer an SCFC-LSTM as the caption generator, which can leverage discriminative semantic information through the caption generation process. Besides, our model is more attractive from the modelling perspective because it can be trained fully end-to-end.

The contributions of this study are as follows:
\begin{enumerate}{}{}
\item{A novel SCFC attention network is proposed. A compounding function is formed, which can perform multistep reasoning on cross-modal features to promote the generation of discriminative semantic features.}
\item{Our multi-aspect features as principal components in the compounding function contain (1) spatial-visual information, (2) high-level semantic concepts, and (3) contextual information to represent various aspects of the given image. Furthermore, there is no need for an independent stage to extract these features since the proposed model can be trained fully end-to-end with a single optimization level.}
\item{We provide SCFC-LSTM as the caption generator, which can use discriminatory semantic information through the caption generation process, thereby generating fine-grained captions.}
\item{We verify the effectiveness of our method on the benchmark MS-COCO and Flickr30K datasets. Experimental results show that the proposed method can achieve competitive results with state-of-the-art methods.}
\end{enumerate}

\section{Related Works}
A large number of articles on image captioning have been published to date. Specifically, we are interested in the visual attention approaches used in the classical encoder-decoder framework, which has attracted considerable interest due to its outstanding performance. Previous works combine a CNN to encode an image into a single static visual feature map and then feed it into an RNN as a decoder [1], [14], [15]. However, this static representation can lead to losing local information.

Inspired by the attention mechanism introduced in machine translation [8], [16] firstly proposed spatial attention in image captioning, which integrates the hidden state of the last step and visual features of patches to calculate the attention weights over different patches in images. A weighted sum of all patches obtains the soft attention feature, and the hard attention feature corresponds to visual information of the most important patch. Although this has made great progress, it involves many meaningless patches, which leads to high computational complexity and increased visual interference. [17] proposed scene-specific contexts and employs selective search [18] to generate region proposals. Besides, a soft attention mechanism [19] proposed an area-based attention mechanism which allows a direct association between caption words and image regions by modelling the dependencies between image regions, caption words, and the hidden states. Furthermore, [20] highlights that the decoder may predict non-visual words with little visual information from the image and introduce a visual sentinel to decide whether to participate in the visual part or language model. Therefore, by introducing adaptive attention, [20] indicates the need to consider processing non-visual words.

Different types of attention mechanisms have been proposed that can use weekly supervised multiple instance learning (MIL) to learn advanced concepts and combine them into sentences to solve the problem of generating non-visual words. [9] introduced a visual attribute classifier to generate semantic concepts in which image features are a vector of attribute classifier confidence. [11] has developed a novel attribute-guided model by integrating inter-attribute correlations into MIL to add the high-level semantic attributes into an RNN-based encoder-decoder framework to achieve better performance. They have constructed various architectures to feed these features to find the best way to incorporate semantic attributes. However, all attribute words are considered equally essential and incorporated into the RNN at every time step. Involving all of the probable words in the image may lead to generating some unrelated words to the image context in the final caption.

All in all, existing methods have tended to focus on either visual or attribute information. Due to their complementarity, they lead to insufficient knowledge of the given image. The other problem is that a few works consider the importance of preserving contextual information in the caption generation process. The generation of a subjective sentence to point out the salient event is strongly influenced by the scene in which the image appears.

\section{The Proposed Method}
\begin{figure*}[!t]
\centering
\includegraphics[width=0.8\textwidth]{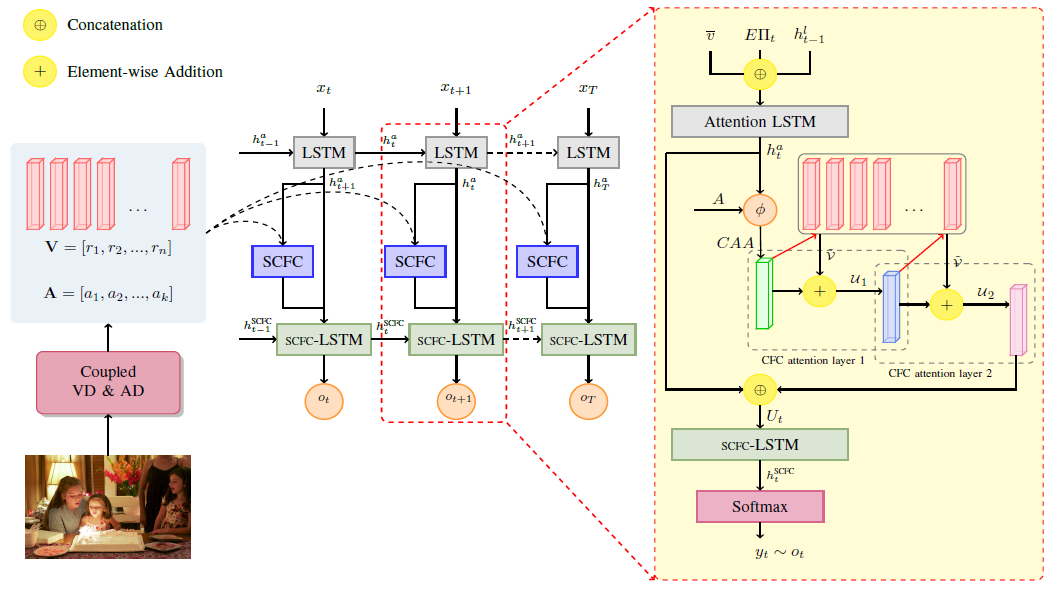}
\caption{The overview of the proposed SCFC for image captioning. Region proposals and attributes are extracted at the first step and then fed to the SCFC cell in each time step to consolidate cross-modal features through the caption generation process.}
\label{fig_1}
\end{figure*}

This section introduces details of the proposed SCFC attention network for image captioning. The overview of the proposed framework is illustrated in Fig.\ref{fig_1}. The framework comprises three components: (1) the coupled visual detector and attributes predictor, (2) the SCFC attention network, and (3) captioning with SCFC-LSTM. First, we extract visual regions \(V=\left\{v_{1}, \ldots, v_{n}\right\}, v_{i} \in \mathbb{R}^{h}, V\in\mathbb{R}^{n \times h}\) and semantic attributes \(A=\left\{a_{1}, \ldots, a_{c}\right\}, a_{i} \in \mathbb{R}^{|\Sigma|}, A\in\mathbb{R}^{|\Sigma|\times c}\) from the given image in a coupled manner. Then, they are fed to the SCFC module. After constructing cross-modal features as principal components, the proposed CFC attention network is triggered to form a compounding function inspired by the multi-step reasoning idea. Finally, the consolidated semantic features forward to the SCFC-LSTM to generate semantically fine-grained image captions. 

\subsection{Coupled Visual Detector and Attribute Predictor}
\begin{figure*}[!t]
\centering
\includegraphics[width=\textwidth]{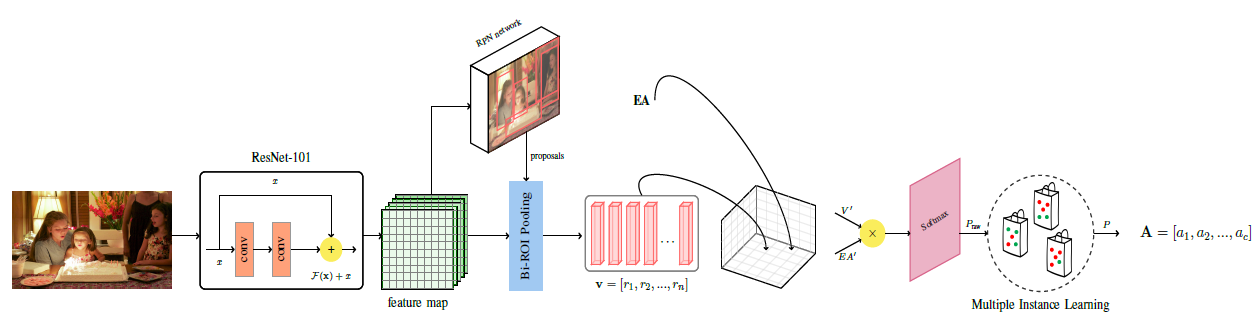}
\caption{The architecture of the coupled VD and AP in our approach. Given an image, the figure shows the process of detecting visual regions V and attributes A.}
\label{fig_2}
\end{figure*}

Visual and semantic features are complementary to each other. With this in mind, we develop a model that can leverage both to enhance the generation of visual and non-visual words like “helping” and “sitting”. In traditional works, predicting attributes is treated as an independent task and depends on a standalone stage, increasing the overall number of model parameters. Inspired by the end-to-end attribute detection in [21], we adopt an attribute predictor (AP) that can be trained jointly with the whole captioning network. Different from previous studies [21] [22], our visual detector (VD) can also be trained with the entire captioning network. In particular, we argue that training feature extractors within the whole system can boost the model to extract more task context-related features. We first detect a set of salient regions of the image, and then on the top of the visual detector, we construct our attribute predictor. The architecture of the coupled VD and AP is given in Fig. \ref{fig_2}.

\subsubsection{Visual Detector}
To extract the deep information of the salient candidate regions, we use ResNet-101 [23] architecture to obtain feature maps of the input image. We employ a modified region proposal network (RPN) [24] to detect regions in the given image with a set of rectangular region proposals and corresponding confidence scores.

In the proposed architecture, after the final convolutional layer of ResNet-101 [23], a \(3\times3\) sliding window moves across the feature map and maps it to a lower dimension (e.g., 256-d). Multiple possible regions based on \(k\) fixed-ratio translation-invariant anchor boxes are generated for each  \(3\times3\) window of the feature map. Thus, the regression layer generates \(4k\) output representing the bounding boxes of the regions, and the classification layer generates \(2k\) outputs representing the softmax probability of each of the \(k\) bounding boxes as a confidence score. We set the value of \(k\) as \(9\), which includes \(3\) scales and \(3\) aspect ratios for each scale. Finally, the bilinear interpolation [25, 26] is used to enhance the nearest neighbour interpolation method in the original ROI pooling layer in [24] so that our model can extract a fixed-sized representation \(V=\left\{v_{1}, \ldots, v_{n}\right\}, v_{i} \in \mathbb{R}^{h}, V\in\mathbb{R}^{n \times h}\) smoothly from each region. Moreover, bilinear interpolation allows end-to-end backpropagation through the region proposals.

\subsubsection{Attribute Predictor}
\begin{figure}[!t]
\centering
\includegraphics[width=0.9\columnwidth]{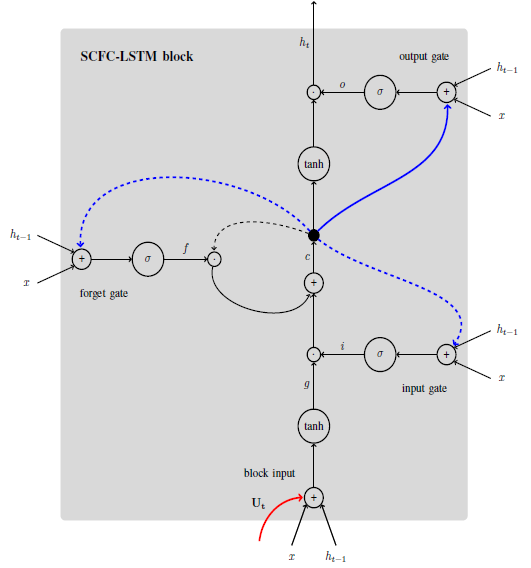}
\caption{The structure of our proposed SCFC-LSTM. Peephole connections and our consolidated input are shown with red and blue lines, respectively.}
\label{fig_3}
\end{figure}

The proposed AP uses the extracted salient regions and the attribute embedding EA generated by embedding LSTM as a by-product to model the similarity between object features and attributes during the detection process. As shown in Fig. \ref{fig_3}, the probability that a given image contains an attribute is predicted in two steps. 

In the first step, we map the attribute embedding and the object features to the same space using two fully connected layers. Then, these are combined using matrix multiplication to measure the similarity. The output of these steps feeds to a softmax layer to generate the raw probability matrix \(P_{raw} \in \mathbb{R}^{1000 \times k}\), where \(P_{raw}^{ij}\) stands the raw probability that the \(j^{th}\) region contains the \(i^{th}\) attribute \(a_{i}\). \(P_{raw}\) is obtained based on Eq. \ref{eq:eq1}.

\begin{equation}
P_{\text {raw}}=\operatorname{sigmoid}\left(\left(W_{\text {AP}} EA\right)^{T} \otimes W_{v} V^{T}\right)
\label{eq:eq1}
\end{equation}

\noindent where \(W_{AP}\in \mathbb{R}^{d\times e}\) and \(W_{v} \in \mathbb{R}^{d\times h}\) are trainable parameters. \(E\in \mathbb{R}^{e\times |\Sigma|}\) represents the embedding of all the words in the vocabulary \(\Sigma\) with embedding size \(e\),  \(A\in \mathbb{R}^{|\Sigma|\times c}\) is the one-hot index matrix of the \(c\) attributes, \(\otimes\) denotes the matrix multiplication, and the superscript \(T\) is the transpose operation.

In the second step, the probability values in each row of \(P_{raw}\) are combined using the noisy-OR multiple instance learning (MIL) method [27] to generate the final probability \(P_{i}\) (Eq. \ref{eq:eq2}) that the input image contains the \(i^{th}\) attribute \(a_{i}\).

\begin{equation}
p_{i}=1-\prod_{j=0}^{n}\left(1-P_{r a w}^{i j}\right)
\label{eq:eq2}
\end{equation}

We face two imbalanced training set problems for training the coupled visual detector and attribute predictor. The number of regions proposed in the RPN network could be as high as several hundred thousand, most of which are negative examples since there is no object inside. Only a fixed number of samples with a fixed object/not-object score is sampled in classical training to overcome the class imbalance problem. Besides, the ground truth attribute vectors are sparse, as a few attributes appear in the ground truth captions.

Focal loss [28] (Eq. \ref{eq:eq3}) is leveraged to defeat this problem, in which all pre-located concrete anchors are taken for training with a dynamically cross-entropy loss. Although all anchors are considered, overwhelming the detector is prevented by weighting the losses of easy samples. Likewise, this modification applies to the attribute predictor, where we treat the non-existent attributes in the ground truth captions as negative examples in RPN.

\begin{equation}
FL(p)=  
\begin{cases}
  -\alpha (1-p)^{\gamma}\log(p),  & \text{y=1}
  \\
  -(1-\alpha) p^{\gamma}\log(1-p), & \text{otherwise}
\end{cases}
\label{eq:eq3}
\end{equation}

We define the visual detector loss \(\mathcal{L}_{VD}\) and attribute predictor loss \(\mathcal{L}_{AP}\) as Eq. \ref{eq:eq4} and \ref{eq:eq5}, where \(\mathcal{L}_{reg}\) is the smooth \(\mathcal{L}1\)  loss used in the regression layer. \(t_{i}^{*}\) and \(p_{i}^{*}\) are the regression target and object/not-object labels. \(\lambda\) is a balancing weight, \(N_{cls}\) and \(N_{reg}\) are the normalization terms, and \(N_{pos}\) is the number of positive attributes.

\begin{equation}
\mathcal{L}_{VD}=\frac{1}{N_{cls}}\sum_{i}FL(p_{i})+\lambda\frac{1}{N_{reg}}\sum_{i}p_{i}^{*}\mathcal{L}_{reg}(t_{i}, t_{i}^{*})
\label{eq:eq4}
\end{equation}

\begin{equation}
\mathcal{L}_{AP}=\frac{1}{N_{pos}}\sum_{j=1}^{c}FL(p_{j})
\label{eq:eq5}
\end{equation}

It should be noted that \(y=1\) in Eq. \ref{eq:eq3} means the anchor contains an object, and the attribute \(a_{j}\) exists in the ground truth captions when we calculate \(\mathcal{L}_{VD}\) and \(\mathcal{L}_{AP}\) losses, respectively. Further,  \(\alpha\), \(\gamma\) and \(\lambda\) hyper-parameters are empirically set to \(0.3|0.95\), \(2|2\), and \(10|-\) for visual detector and attribute predictor losses, respectively. Finally, the \(\mathcal{L}_{VDAP}\) loss is calculated using Eq. \ref{eq:eq6}.

\begin{equation}
\mathcal{L}_{VDAP}=\mathcal{L}_{VD}+0.5\mathcal{L}_{AP}
\label{eq:eq6}
\end{equation}

\subsection{SCFC Attention Networks}
At the heart of our proposed method is a compounding function to consolidate cross-modal features so that we can guide the language model in the caption generation process. Our compounding function aims to combine cross-modal features in a multi-step fashion and enable the model to attend to the constructed discriminative features in generating all semantic-level words. In the initial step, we form our principal components to feed into the proposed compounding function, which comes from two different modalities: textual \(\mathcal{H}\) and visual \(\mathcal{V}\). Then, we define our novel recursive function to compound cross-modal features through the attention networks.

\subsubsection{The Principal Components}
We define our principal components from two modalities: textual \(\mathcal{H}\) and visual \(\mathcal{V}\). The visual element is provided from the output of the visual detector. In this case, we have a set of regional feature maps to participate in the compounding function. As the second component, we leverage semantic attributes regarding the contextual environment. Previous studies employ semantic attention to involve high-level semantic concepts in the caption generation process. The kernel of semantic attention-based methods drives the model to dynamically attend to semantically essential attributes in each time step regarding the contextual information. This core has been formed by learning an attention activation state vector to calculate the weight of each attribute. Existing methods meet this goal by adding elementwise the hidden state vector of the language LSTM from the previous time step and each attribute vector. Then, the weight of each is computed through the softmax layer, and the “soft” approach is followed to obtain the output attention by using the weighted sum of the detected attributes. Note that the output attention vector may contain irrelevant attributes, making the attention guidance vague. We provide two toy examples inspired by a comprehensive investigation conducted in [12] to find the best function for measuring the similarity in considering extreme cases to make sense better.

In the first case, consider the attribute vector A as an all-one vector, and the hidden state of the language LSTM \(h\) is an all-zero vector expressing there is no contextual information. As \(h\) is an all-zero vector, there is no correlation between the attribute vector and the context. Hence, the attention activation state vector should be an all-zero vector meaning no association between the context and the attribute vector. In the second case, imagine the attribute vector A is an all-zero, and the hidden state of the language LSTM \(h\) is an all-one vector, which indicates there are no high-level concepts in the given image. Thus, the corresponding attention activation state vector is supposed to be an all-zero vector, which shows no correlation between the context and the attribute vector. Although the expected result in both examples is an all-zero vector, the attention activation state vector taken by the traditional semantic attention mechanism is an all-one vector leading to the inaccurate weights added to the detected attributes. From the examples, we can find that in some cases, the "soft" attention mechanism may lack an appropriate measure of the correlation between contextual information and high-level semantic concepts. The Context-Aware Attribute CAA suggested in this paper seeks to address this issue by formulating a function measuring the correlation between the predicted attributes and contextual information. Besides, we use the Attention LSTM [22] to represent the contextual information rather than take the hidden state of the language LSTM leading to a proper dynamic representation of the image context condition in the current linguistic context.

As illustrated in Eq. \ref{eq:eq7}, attention LSTM takes the mean pooled image feature \(\bar{v}=\frac{1}{N}\sum_{i}v_{i}\), the previous hidden state of the SCFC LSTM and encoding of the previously generated word as the inputs. These inputs provide the attention LSTM with maximum context regarding the state of the language LSTM, the overall content of the image, and the partial caption output generated so far, respectively.

\begin{equation}
h_{t}^{att}=LSTM[(h_{t-1}^{SCFC}\oplus\bar{v}\oplus E\Pi_{t}), h_{t-1}^{att}]
\label{eq:eq7}
\end{equation}

\noindent where \(E\) stands for the word embedding matrix shared with the attribute detector module, and \(\Pi_{t}\) is the one-hot encoding of the input word at time step \(t\). Further, we choose linear functions to measure the correlation between the predicted attributes and contextual information as low-cost functions in terms of computational efficiency and simplicity. Due to our refinement purpose, we define \(\Phi_{cr}\) as the element-wise multiplication. Thus, the semantic attention distribution \(\beta_{t}^{j}\) over each attribute \(a_{t}^{j}\) can be calculated using Eq. \ref{eq:eq8} and \ref{eq:eq9}.

\begin{equation}
b_{t}^{j}=\Phi_{cr}(h_{t}^{att}, EA_{t}^{j})
\label{eq:eq8}
\end{equation}

\begin{equation}
\beta_{t}^{j}=softmax(b_{t}^{j})
\label{eq:eq9}
\end{equation}

\noindent where \(b_{t}^{j}\) denotes the attention activation state vector of each attribute. Then, we can construct the context-aware attributes \(CAA_{t}\) by multiplying the embedding of each attribute by its weight based on Eq. \ref{eq:eq10}.

\begin{equation}
CAA_{t}=\sum_{j=1}^{c}(\beta_{t}^{j}\cdot EA_{t}^{j})
\label{eq:eq10}
\end{equation}

It should be noted that the proposed CAA vector is supposed to represent those attributes that are associated with contextual information. Intuitively, the coexistence of the final attribute set can preserve the context dynamically in each time step through the caption generation process.

In summary, a set of regional feature maps, semantically embedding of the attributes, and contextual information are all considered to form our principal components of the compounding function. From this information, we provide visual element \(V=\left\{v_{1}, \ldots, v_{n}\right\}, v_{i} \in \mathbb{R}^{h}, V\in\mathbb{R}^{n \times h}\) as the first component, and our suggested context-aware attributes CAA as the second component \(\mathcal{H}\) coming from textual modality.

\subsubsection{Compounding Function}
Given the principal components \(\mathcal{V}\) and \(\mathcal{H}\), we propose a new recursive function to consolidate cross-modal features in a multi-step fashion. In particular, we define an SCFC Attention Network to perform the consolidation. We then set up a stacked network with more SCFCs to operate together to simulate multi-step reasoning. Firstly, we introduce CFC, a fully functional operating standalone, and in the rest of this section, we look at how the stacked network works as a recursive compounding function.

In the SCFC, we first measure the inter-modality relations between principal component pairs to determine the relevance degree of the cross-modal features. Like the previous part, we use \(\Phi_{cr}\) for calculating the relevance degree as below:

\begin{equation}
\alpha_{i,t}=\tanh(\Phi_{cr}(W_{\mathcal{V},\alpha}v_{i},W_{\mathcal{H},\alpha}\mathcal{H}_{t}))
\label{eq:eq11}
\end{equation}

\begin{equation}
\mathcal{D}_{i,t}=softmax(W_{\alpha,\mathcal{D}}^{T} \alpha_{i,t})
\label{eq:eq12}
\end{equation}

\noindent where \(W_{\mathcal{V},\alpha} \in \mathbb{R}^{d\times h}\), \(W_{\mathcal{H},\alpha} \in \mathbb{R}^{d\times e}\) and \(W_{\alpha,\mathcal{D}}^{T} \in \mathbb{R}^{h}\) are learned parameters. By determining the relevance degree \(\mathcal{D}_{i,t}\) between textual component \(\mathcal{H}\) and visual sub-components \(v_{i}\), we calculate attended visual component \(\mathcal{V}\) as follows:

\begin{equation}
\Tilde{\mathcal{V}}_{t}^{I}=\sum_{i}\mathcal{D}_{i,t}\mathcal{V}_{i}
\label{eq:eq13}
\end{equation}

The calculated \(\Tilde{\mathcal{V}}_{t}^{I}\) represents the visual element, which is strained with the textual component at each time step \(t\). To solidify the impact of the textual information, we define the compact cross-modal representation \(\mathcal{U}_t\) as a second-order combination as follows:

\begin{equation}
\mathcal{U}_t=\Tilde{\mathcal{V}}_{t}^{I}+\mathcal{H}_{t}
\label{eq:eq14}
\end{equation}

The output \(\mathcal{U}_t\) is an informative representation in which the principal components \(\mathcal{V}\) and \(\mathcal{H}\) of a given image are encoded. Compared with the models that only adopt combined visual and semantic attention, our model constructs a richer \(\mathcal{U}_t\) by imposing higher weights on the visual elements that are more relevant to the context-aware attributes. However, for abstract semantic words, a single CFC is insufficient for making inferences on the cross-modal features and generating an all-round representation. Therefore, we stacked SCFCs to iterate the above procedure imitating the multi-step reasoning approach in multi-modal tasks. Mathematically, for the \(s^{th}\) CFC layer, the SCFC takes the following formulas:

\begin{equation}
\alpha_{i,t}^{s}=\tanh(\Phi_{cr}(W_{\mathcal{V},\alpha}^{s}v_{i},W_{\mathcal{H},\alpha}^{s}\mathcal{H}_{t}^{s-1}))
\label{eq:eq15}
\end{equation}

\begin{equation}
\mathcal{D}_{i,t}^{s}=softmax(W_{\alpha,\mathcal{D}}^{s,T} \alpha_{i,t}^{s})
\label{eq:eq16}
\end{equation}

\noindent where \(\mathcal{H}_{t}^{s-1}\) is the textual element from the previous SCFC layer at each time step \(t\). It should be noted that we initialize \(H_{t}^{0}\) with the context-aware attributes vector. In each reasoning step, the attended visual component \(\mathcal{V}\) and the cross-modal representation \(\mathcal{U}_t\) are obtained as below:

\begin{equation}
\Tilde{\mathcal{V}}_{t}^{I,s}=\sum_{i}\mathcal{D}_{i,t}^{s}\mathcal{V}_{i}
\label{eq:eq17}
\end{equation}

\begin{equation}
\mathcal{U}_t^s=\Tilde{\mathcal{V}}_{t}^{I,s}+\mathcal{H}_{t}^{s-1}
\label{eq:eq18}
\end{equation}

We repeat this algorithm \(S\) times and then use the final \(\mathcal{U}_t^S\) concatenated with the contextual representation \(h_{t}^{att}\) as Eq. \ref{eq:eq7} to serve it to the SCFC-LSTM to generate semantically fine-grained sentences.

\begin{equation}
U_t=\mathcal{U}_{t}^{S}\oplus h_{t}^{att}
\label{eq:eq19}
\end{equation}

\subsection{Captioning with SCFC-LSTM}
There are several LSTM variants. In our work, we adopt the peephole LSTM [29, 30] model as our caption generator, which is expressed as follows:

\begin{equation}
\begin{aligned}
i_t & =\sigma(W_ix_t+R_{i}h_{t-1}+P_{i}c_{t-1}+b_i)\\
f_t & =\sigma(W_fx_t+R_{f}h_{t-1}+P_{f}c_{t-1}+b_f)\\
o_t & =\sigma(W_ox_t+R_{o}h_{t-1}+P_{o}c_{t}+b_o)\\
g_t & =\tanh(W_gx_t+R_{g}h_{t-1}+b_g)\\
c_t & =f_t\odot c_{t-1}+i_t\odot g_t\\ 
h_t & =o_t\odot \tanh(c_{t})
\end{aligned}
\label{eq:eq20}
\end{equation}

\noindent where \(i_t\), \(f_t\), \(o_t\), \(c_t\) and \(h_t\) are the input gate, forget gate, output gate, memory cell, and hidden state of the peephole LSTM, respectively. \(\sigma\) indicates the sigmoid function, \(x_t\) shows the word input at the time step \(t\). \(P_* \in \mathbb{R}^N\), \(W_*\in \mathbb{R}^{N\times M}\), \(R_*\in\mathbb{R}^{N\times N}\), and \(b_*\in \mathbb{R}^N\) denote peephole, input, recurrent, and bias weights, respectively, where \(N\) is the number of LSTM blocks and \(M\) is the dimension of inputs. It is worth mentioning that adding peephole connections means that we let the gate layers look at the cell state. In other words, when determining input gates, forget gates and output gates, there is a need to utilize the previous time step of the cell state \(c_t\).

Previous studies utilize semantic features associated with either particular input locations or high-level semantic concepts as additional inputs of the language model at each time step. On the contrary, our model provides the cross-modal consolidated feature \(U\) to guide the description generation. As shown in Fig. \ref{fig_1}, we load the constructed feature \(U\) into cell state \(C\) at each time step. Intuitively, peephole connections also allow the current time step of the gate to be aware of cross-modal informative semantic feature \(U\) in a more governable way. Thus, the fourth line in Eq. \ref{eq:eq20} must be updated as below:

\begin{equation}
g_t =\tanh(W_gx_t+W_{U}U_{t}+R_{g}h_{t-1}+b_g)\\
\label{eq:eq21}
\end{equation}

\noindent where \(W_U\) indicate a weight matrix. Finally, the probability distribution over each word in the vocabulary \(p_t\) and the word to be generated at time step \(t\) is predicted as:

\begin{equation}
h_{t}^{SCFC}=LSTM^{P}(h_{t-1}^{SCFC},x_t,\mathcal{U}_{t}) 
\label{eq:eq22}
\end{equation}

\begin{equation}
y_t\sim p_t=softmax(W^{h}h_{t}^{SCFC}) 
\label{eq:eq23}
\end{equation}

\noindent where \(W\) indicates a weight matrix, and \(LSTM^{P}\) denotes the LSTM with peephole connections.

\subsection{Model Training}
The model training process consists of two rounds. In the first round, given a target ground truth sequence \(y_{1:T}^{*}\), we optimize the model with the classical cross-entropy (XE) loss as Eq. \ref{eq:eq24}, where \(\theta\) stands for the captioning model parameters.

\begin{equation}
\mathcal{L}_{XE}(\theta)=-\sum_{t=1}^{T}\log(p_\theta (y_{t}^{*}|y_{1:t-1}^{*}))
\label{eq:eq24}
\end{equation}

In the second round, we leverage deep reinforcement learning (RL) to address the exposure bias problem, which means the model has never been exposed to its predictions, resulting in accumulated errors during the inference process. From the initialization of the model trained by cross-entropy, we investigate to minimize the negative expected score corresponding to the model parameter as below:

\begin{equation}
\mathcal{L}_{R}(\theta)=-E_{y_{1:T}\sim p_\theta}[r(y_{1:T})]
\label{eq:eq25}
\end{equation}

\noindent where \(r\) is the evaluation score metric optimized with the CIDEr-D [31] score. Considering the self-critical sequence training (SCST) [32], we approximate the gradient by the REINFORCE algorithm, given by:

\begin{equation}
\nabla\mathcal{L}_{R}(\theta) \approx -(r(y_{1:T}^{s})-r(\hat{y}_{1:T}))\nabla_{\theta}\log p_{\theta}(y_{1:T}^{s})
\label{eq:eq26}
\end{equation}

\noindent where \(y_{1:T}^{s}\) is a randomly sampled caption, and \(r(\hat{y}_{1:T})\) is the baseline score of the max sampled caption.

Our proposed model is trained using a single-level optimization. Specifically, the model is optimized with the overall loss function \(\mathcal{L}_{o}\), as shown in Eq. \ref{eq:eq27}. \(\mathcal{L}_{VDAP}\) is defined as Eq. \ref{eq:eq6}, and \(\mathcal{L}_{cap}\) is cross-entropy and reinforcement learning losses in the first and second rounds, respectively.

\begin{equation}
\mathcal{L}_{o}=\mathcal{L}_{cap}+\mathcal{L}_{VDAP}
\label{eq:eq27}
\end{equation}

\section{Experiments}
\subsection{Datasets}
We validate the proposed model on well-known datasets, including MSCOCO [33] and Flicker30k [34]. MSCOCO contains 82,783 images for training and 40,504 images for validation. Each image has five annotated sentences. We employ the Karpathy splits [1], used widely for reporting results in prior works. This split contains 113,287 training images and 5K images for validation and testing. Flickr30k consists of 31,783 images obtained from Flickr. For a fair comparison, we use the publicly split [1] with 29,783 images for training, 1K for validation, and 1K for testing. Like MSCOCO, each image is annotated with five reference captions.

\subsection{Evaluation Metrics}
We report the performances with the popular metrics for image captioning, including BLEU-N [35], METEOR [36], CIDEr [31], ROUGE-L [37], and SPICE [38]. We use the code published by the Microsoft COCO evaluation server to calculate all metrics. BLEU is computed by measuring the similarity of the generated sentences and the reference sentences in n-grams. METEOR is evaluated by comparing the various segments of sentences between the candidate caption and the reference caption. CIDEr is a consensus-based metric introduced specifically for image captioning tasks, which calculates consensus in image description by performing TFIDF weighting for all n-grams. ROUGE-L is used to evaluate the adequacy and fluency of machine translation, which employs the longest common subsequence between a candidate sentence and a set of reference sentences to measure their similarity at the sentence level. SPICE is determined by employing an F-score measured over tuples in the reference and candidate scene graphs, obtained through dependency parse trees.

\subsection{Experimental Settings}
\subsubsection{Data Preprocessing}
We adopt the standard practice and apply only minimal text preprocessing, tokenizing on white space, converting all tokens to lowercase, and discarding rare words that occur fewer than five times, resulting in the remaining words in the vocabulary of sizes 10,010 and 6,864 for MSCOCO and Flickr30k respectively. In particular, we replace less frequently occurring words with a special token \(<UNK>\).

\subsubsection{Implementation Details}
Our ground truth includes positive boxes and attributes vectors where coordinate annotations are obtained from the MSCOCO-2014 object detection dataset. To build the ground truth attribute, we select \(c\) most common words in the training caption corpus, where we set \(c=1000\) as in [39]. Then, we construct an attribute vector to determine whether each word exists in each image’s description. We employ ResNet-101 with weights pre-trained on ImageNet [40] without fine-tuning in all the training phases. Our visual detector extracts regional features with a size of \(36\times2048\)-d according to the corresponding confidence scores as suggested in [22]. Similar to [21], attributes share the same embedding with the corresponding words, where each word is embedded in a \(1000\) dimensional word embedding space. The dimensions of both attention LSTM and SCFCLSTM hidden states are set to \(2048\)-d. The Adam optimizer [41] with \(\beta_{2}=0.9\), \(\beta_{2}=0.999\) is utilized to minimize the loss function in Eq. \ref{eq:eq26}. The basic learning rate is \(1\times10^{-4}\). Dropout [42] and gradient clipping techniques [43] are used. In the first round of training, we adopt teacher-forced learning [44], in which we provide the ground truth words up to \(t-1\), and not the words it generated in the previous time step to train the prediction at \(t\) for sequence learning. In the testing phase, the maximum allowable sentence length is set to \(20\). We use the beam search strategy, and the beam size is set to \(3\).

\subsubsection{State-of-the-art Studies}
We confirm the effectiveness of our method by comparing its performance with the following state-of-the-art works:

\begin{enumerate}{}{}
\item{NIC [14]: The first encoder-decoder framework that takes an image as input in an encoder and feeds the encoded representation into the first time step of the LSTM-based decoder to generate the corresponding description.}
\item{Soft-Att [8] and Hard-Att [8]: Two different spatial attention mechanisms are introduced to guide the model to selectively attend to the salient image regions in either deterministic “Soft” attention or stochastic “Hard” attention.}
\item{Sem-Att [9]: First, the semantic concepts are detected in the image as an independent task. Then, the global image features and the detected semantic concepts are combined, and they are progressively fed into the language model through the caption generation process.}
\item{LSTM-A [11]: They have suggested a novel attribute detector integrating inter-attribute correlations into multiple-instance learning (MIL) to leverage correlations between attributes. Then, they use five different forms to combine those semantically attributes into the LSTM.}
\item{SCA-CNN [45]: This is an improved version of visual attention that incorporates spatial, channel, and multi-layer image features to dynamically adjust the context of sentence generation.}
\item{SCST [32]: An advanced reinforcement learning (RL) based training method is proposed for image captioning.}
\item{Ada-Att [20]: Adaptive attention with a visual sentinel is proposed to determine whether to attend to the visual features or the visual sentinel.}
\item{RFNet [46]: They have proposed a novel recurrent fusion network (RFNet) that can exploit the interactions between the outputs of the image encoders and generate new compact and informational representations for the decoder.}
\item{Up-Down [22]: A combined top-down and bottom-up attention mechanism is introduced that enables attention to be determined at the level of objects and other salient image regions.}
\item{GCN-LSTM [47]: They have built a graph convolutional network over the detected objects in an image that integrates semantic and spatial object relationships into the image encoder in an encoder-decoder framework.}
\item{ARL [48]: This work considers a solution to integrate intra-regional relationships into visual content and proposes a novel visual attention mechanism to implicitly model the relationships between image regions.}
\item{SGAE [49]: A scene graph auto-encoder has been proposed that incorporates collocations and contextual inference into the encoder-decoder architecture as the language inductive bias by using the scene graph of the image and a trained dictionary.}
\item{CAVP [50]: They have proposed a new RL-based learning method and introduced a pairwise relationship learning approach in the decoder.}
\item{LBPF [51]: They have suggested a look back method to embed previously visual information and a predict forward strategy to look into the future to boost image captioning performance by utilizing linguistic coherence.}
\item{MAD+SAP [21]: This work expands semantic attention by introducing a subsequent attribute predictor module to dynamically predict a concise attribute subset at every time step to mitigate the variety of image attributes.}
\end{enumerate}

\subsubsection{Quantitative Results}
Table \ref{tab:table1} illustrates the contrastive performance comparison results on the MS-COCO dataset. The table shows that our proposed model outperforms the state-of-the-art models with a large margin in all evaluation metrics, specifically CIDEr. For a fair comparison, we also separately report the results of the ensemble models. These evaluation results indicate that the SCFC attention network boosts image captioning performance. Compared with the traditional neural image captioner NIC [14], classical visual attention Soft-Att [8], and Hard-Att [8] as benchmarks, our improvement is primarily due to the more effective collaboration of visual and semantic information. Compared with SCA-CNN [45] and LBPF [51], our model uses advanced semantic concepts to generate more diverse and semantic-enriched captions. Although Sem-Att [9] and LSTM-A [11] leverage attributes, our model can more accurately measure the correlation between attributes and contextual information, and there is no need for a separate network to train the attribute detector. Compared with methods that explore visual semantic regions by pre-training a visual detector with a large dataset (such as Visual Genome) before implementing the image captioning network (Up-Down [22], MAD+SAP [21]), we leverage the cross-modal feature consolidation layer to make up for the lack of discriminative semantic feature representation. RFNet [46] further improves image captioning performance by investigating the spatial attention mechanism. 

\begin{table*}[!t]
\caption{Performance analysis of the proposed SCFC and other state-of-the-art methods on the MSCOCO KARPATHY's test split, where B@N, M, R, C and S are the short forms of BLEU-N, METEOR, ROUGE-L, CIDER-D, AND SPICE scores. The most significant number in each column is marked in boldface.\label{tab:table1}}
\centering
\begin{tabular}{c}
\includegraphics[width=1\textwidth]{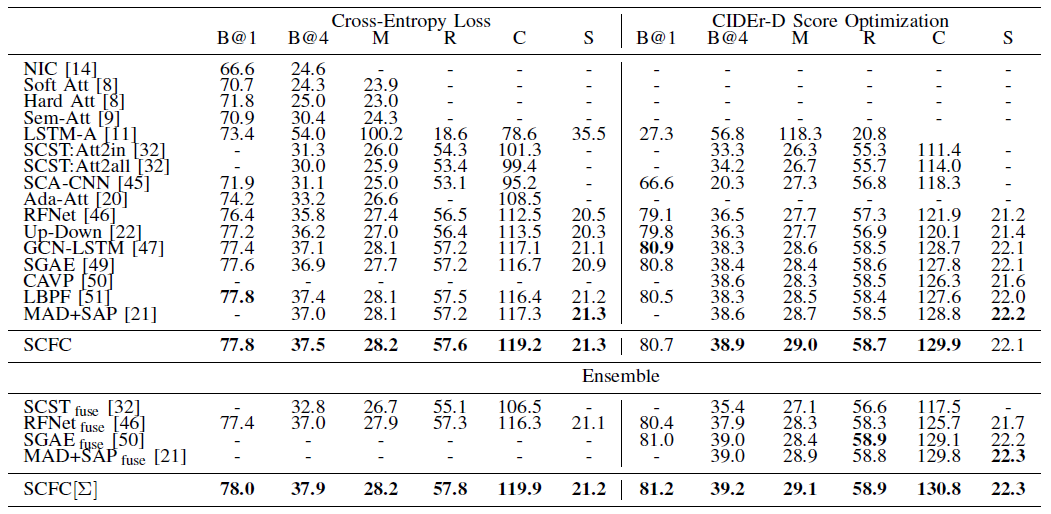}
\end{tabular}
\end{table*}

In contrast, we employ an attention mechanism considering both semantic concepts and spatial regions. ARL [48] and GCN-LSTM [47] consider the visual relationship among regions in the image by discovering the high-level connections between regions that encode semantic concepts, thereby improving image captioning performance. At the same time, our model can capture these regions in compounding functions through the SCFC layers. CAVP [50] proposes a new RL-based learning method employed in our learning procedure to improve image captioning performance. MAD+SAP [21] utilizes visual and semantic information and focuses on the role of attributes. While keeping in mind the complementary nature of visual and semantic information, we propose a fully end-to-end model with a novel consolidation layer to efficiently generate more fine-grained captions. Unlike SGAE [49], which requires pre-training a scene graph generator and a dictionary, our model relies on the visual feature as the only input.

To further verify the effectiveness of the proposed model, we report the evaluation results of the test split of the Flickr30k dataset in Table \ref{tab:table2}. We evaluate the proposed model on the online MSCOCO test server. Table \ref{tab:table3} reports the performance of the SCFC and other state-of-the-art works. The proposed method also achieves the best performance in most metrics.

\begin{table}[!t]
\caption{Performance analysis of the proposed SCFC and other state-of-the-art methods on the Flicker30K publicly split using the cross-entropy loss, where B@N, M, R, C and S are the short forms of BLEU-N, METEOR, ROUGE-L, CIDER-D, AND SPICE scores. The most significant number in each column is marked in boldface.\label{tab:table2}}
\centering
\begin{tabular}{c}
\includegraphics[width=\columnwidth]{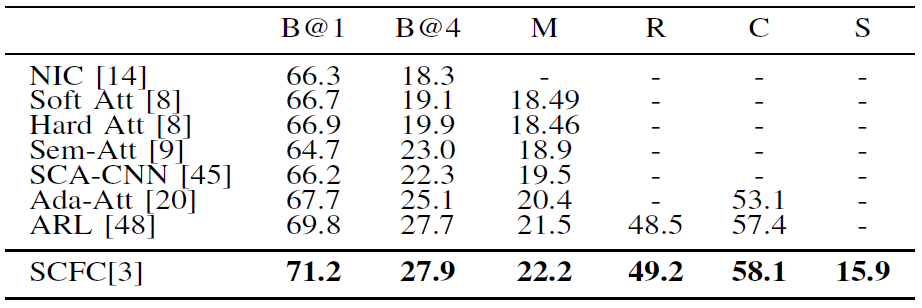}
\end{tabular}
\end{table}

\begin{table*}[!t]
\caption{Performance analysis of the proposed SCFC and other state-of-the-art methods on the online MSCOCO test server. The superscript of each metric indicates the top 2 rankings.\label{tab:table3}}
\centering
\begin{tabular}{c}
\includegraphics[width=1\textwidth]{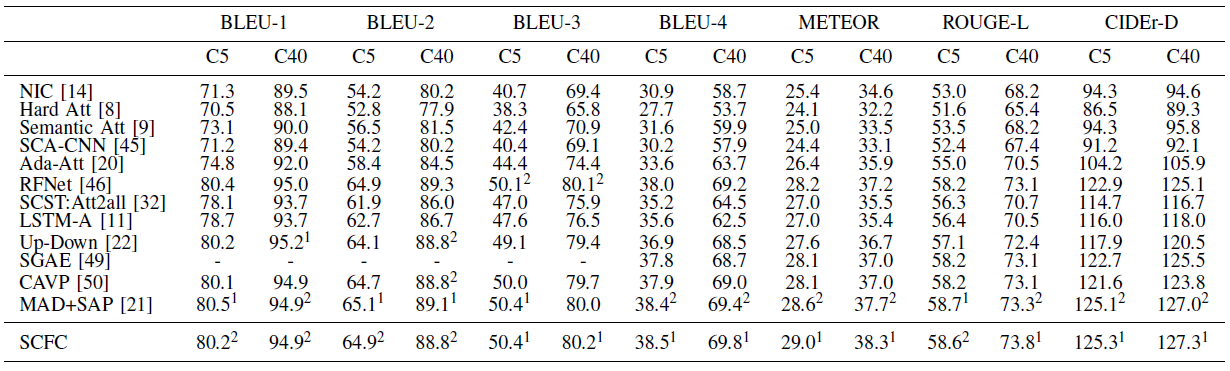}
\end{tabular}
\end{table*}

\subsection{Ablation Studies}
\subsubsection{Incrementally Validation}
We also conduct extensive experiments to incrementally validate our method and thoroughly show the behaviour of the proposed method. We use the following parts to ablate our model:

\begin{itemize}
\item{Base: We construct a baseline model by following [22] to integrate an attention LSTM to the language LSTM coupled with our visual detector. Note that unlike [22], our visual detector can be trained jointly with the whole captioning network.}
\item{Base+VDsemAtt: We inject our attribute detector through the traditional semantic attention as an extra input to the language LSTM.}
\item{Base+CAA: To show the effectiveness of our proposed CAA, we incorporate detected attributes through the context-aware attributes module into the Base model.}
\item{CAA+SCFC: We add the proposed stacked cross-modal feature consolidation attention network to the ablative model Base+CAA. Compared to the entire model, this variant does not adopt SCFC-LSTM.}
\item{CAA+SCFC+SCFC-LSTM: The proposed model with CAA, SCFC attention network, and SCFC-LSTM.}
\end{itemize}

The experimental results of the ablated models are reported in Table \ref{tab:table4}. The number in brackets denotes the number of stacked layers in the SCFC attention network. Our base model achieves comparable performance to [22]. The results of Base+VDsemAtt and Base+CAA demonstrate that using attributes in forming the textual component can provide better results than leveraging visual elements alone (Base). In comparison, our context-aware attributes module improves the model performance significantly. In particular, our ablative model Base+CAA can achieve 37.6 and 124.8 in the BLEU-4 and CIDEr, respectively, making the relative improvement over our baseline model with classical semantic attention by 0.6\% and 2.9\%, respectively. The proposed CAA performs better than traditional semantic attention on the BLEUN metric. Indicating running detected attributes through the caption generation process, regarding how much they are consistent with the contextual environment of the current step, can guide the model to attend to more context-related attributes resulting in meaningful and fluent sentences. The experiment of CAA+SCFC+ SCFC-LSTM(s), which is the full model with a different number of CFC layers, verifies the effectiveness of our stacked cross-modal feature consolidation attention network. In particular, our model CAA+SCFC+ SCFC-LSTM(3) achieves a relative improvement of 0.9\%, 1.9\%, 1.3\%, 1.3\%, 8.0\%, and 0.7\% in terms of BLEU-1, BLEU-4, METEOR, ROUGE-L, CIDEr, and SPICE, respectively. Besides, our model CAA+SCFC+SCFC-LSTM(1) improves results compared with the CAA+SCFC(1) ablative models proofing the positive influence of adopting SCFC-LSTM rather than the standard LSTM.

\begin{table}[!t]
\caption{The results of ablated models (Base, Base+VDsemAtt, BASE+CAA, CAA+SCFC) and our entire model CAA+SCFC+SCFC-LSTM(N) with the different number of CFC layers on the MSCOCO test split in CIDER-D score optimization training in terms of BLEU-N(B@N), METEOR(M), ROUGE-L(R), CIDER-D(C), and SPICE(S).\label{tab:table4}}
\centering
\begin{tabular}{c}
\includegraphics[width=\columnwidth]{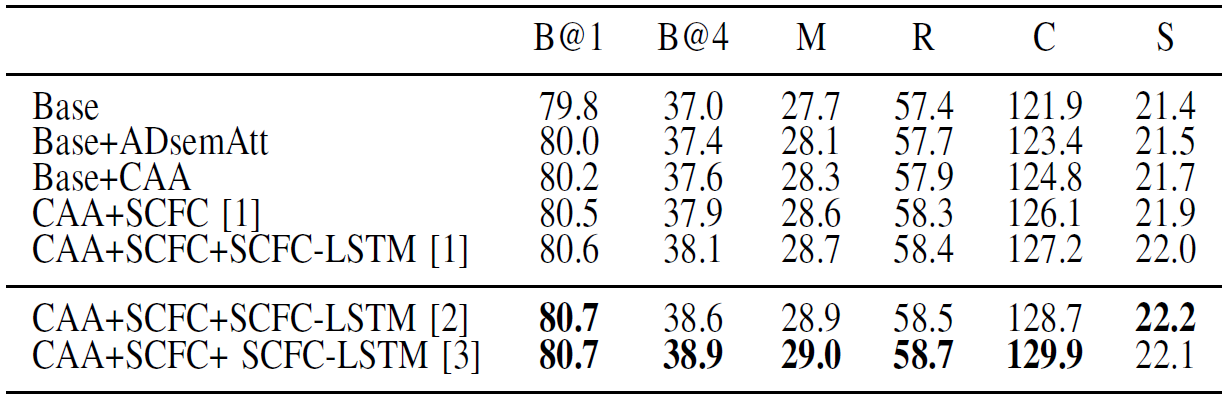}
\end{tabular}
\end{table}

These evaluation results indicate the efficacy of the SCFC layer. By comparing our entire model with its variants, it is not difficult to determine whether the proposed SCFC can combine complementary features to dynamically modulate a richer semantic representation of a given image at each time step. The evaluation results of models with more than one CFC layer further confirm this assertion. The entire model with three stacked CFC layers achieves a relative improvement of 8.5\% in the CIDEr metric, which is remarkable in generating semantic-enriched descriptions.

\subsubsection{Context-Aware Attributes Analysis}
\begin{figure}[!t]
\centering
\includegraphics[width=\columnwidth]{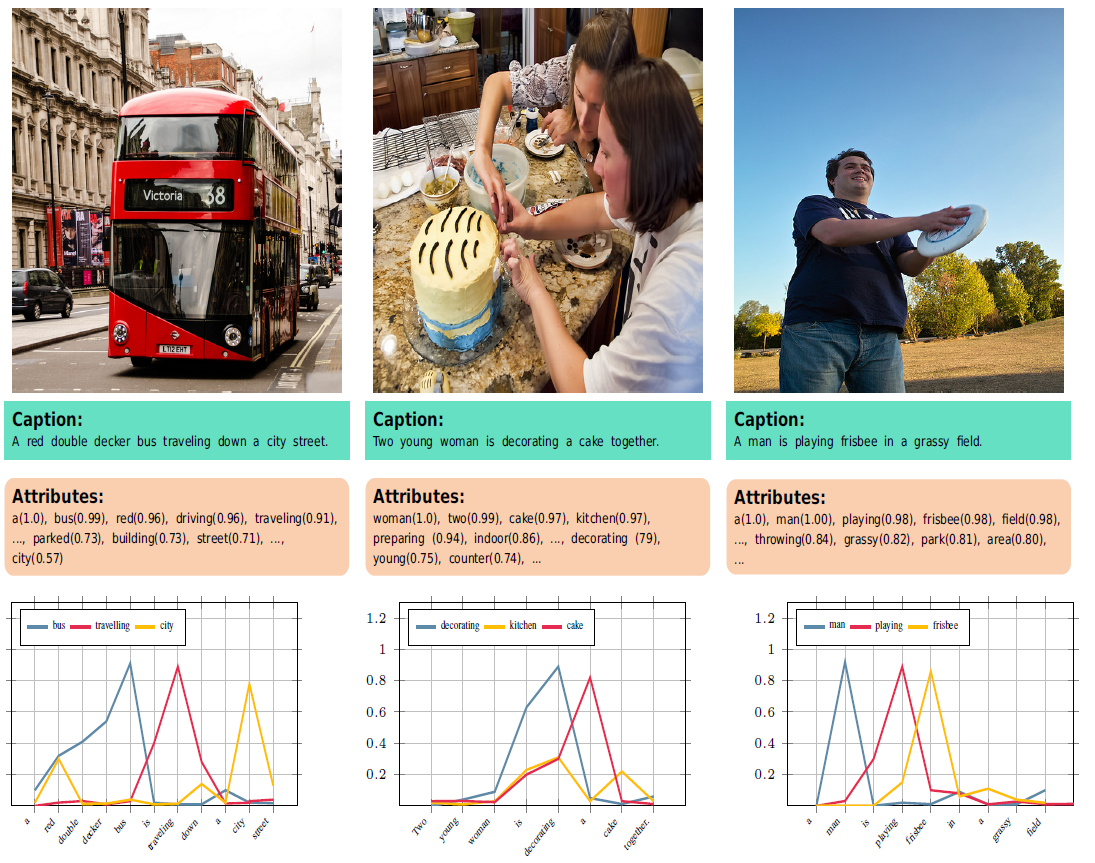}
\caption{Examples of attention weights' changes along with the generation of captions.}
\label{fig_4}
\end{figure}

Fig. \ref{fig_4} shows the impact of semantic attention weights through our proposed context-aware attribute module at each step in our grown model. We only present the attention weights of three probable attributes from the attribute detector output for visual simplicity. While sentences are being generated, variations of the corresponding distinct time-varying weights are adjusted to the current context.

To better understand our model, we preserve three scenarios in CAA weights' changes and attributes' contribution in the generated captions. First, those high probability attributes obtain high weights used in the generated caption. This case almost happens when an attribute can be paired with a well-defined shape region or words describing an action. For example, in Fig. \ref{fig_4}(c), the words "man", "playing", and "frisbee" are picked at the exact time step meaning our CAA can attend accurately to these categories of words. Second, those attributes with a high probability not only obtain high weights but also do not participate in the generated captions. This case takes place when attributes are not consistent with contextual information. For instance, in Fig. \ref{fig_4}(b), although the word "kitchen" is predicted with a high probability (0.97\%), it does not achieve a high weight. Besides, it orients the semantic tendency of the model when words like "decorating" and "cake" are generated, which is the power of the CAA module. Third, those high-probability attributes do not reach high weights through the CAA. Nevertheless, they take a seat in the generated caption. This case shows the ability of the CAA to leverage common catchwords in either spoken or written language. The term "city street" in Fig. \ref{fig_4}(a) is a good example indicating the attention to the word "city", although almost a low probability predicts it.

\subsubsection{The Role of SCFC Attention Layers}
Regarding the generation of visual and non-visual words, feature compounding is beneficial to image captioning. Comparing sentences generated by the SCFC attention network and those generated by the Base model reveals this assertion. In particular, the compounding function in the SCFC layers aims to bring together semantic attributes and visual regions, leading to the generation of fine-grained captions.

\begin{figure*}[!t]
\centering
\includegraphics[width=0.8\textwidth]{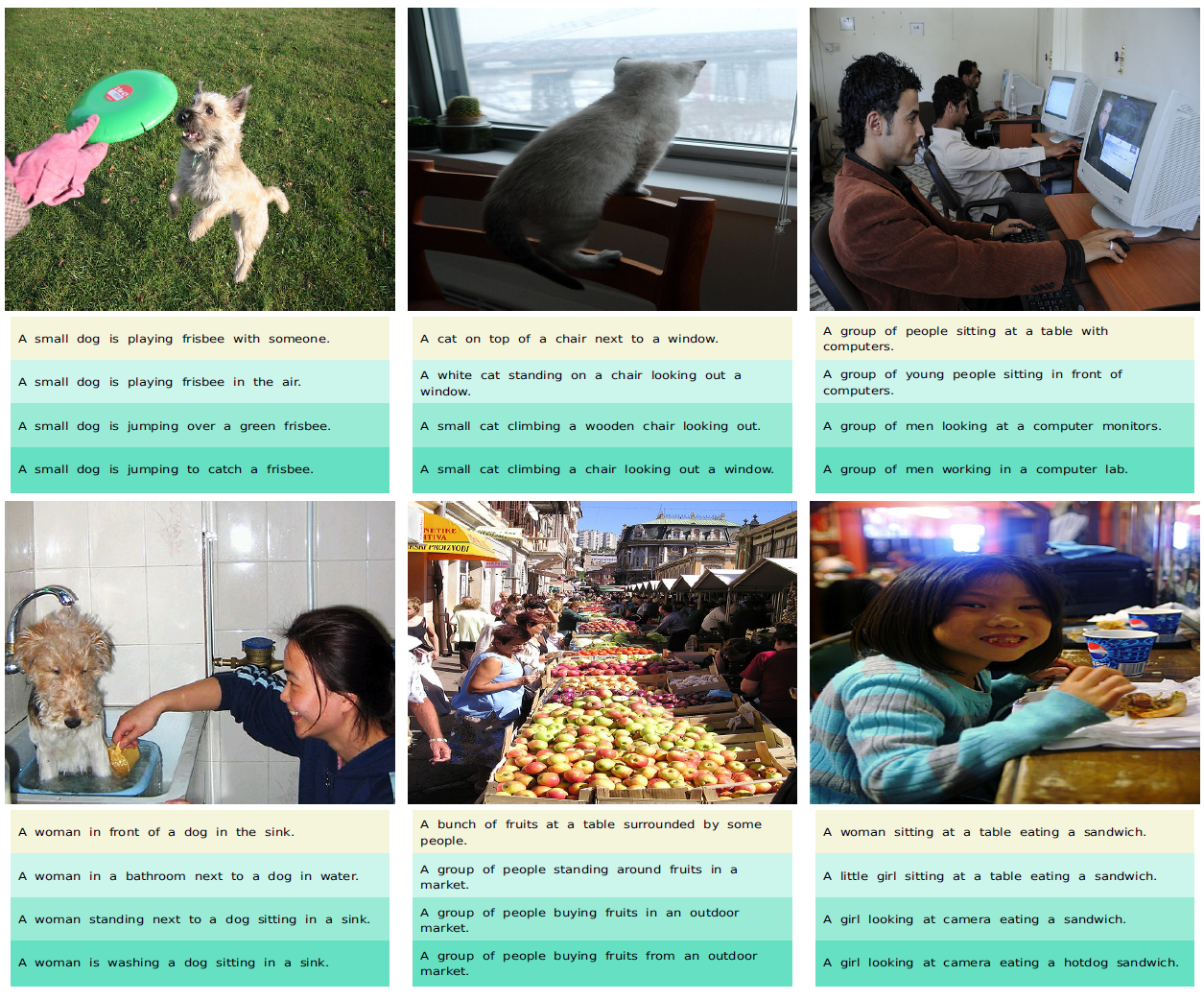}
\caption{Examples of caption results. The captions are generated by the Base model, SCFC, with one/two/three CFC layers in the yellow box to the solid green box, respectively.}
\label{fig_5}
\end{figure*}

We compare the result captions generated by the various number of CFC layers against the baseline model, as shown in Fig. \ref{fig_5}. The figure shows that our SCFC attention network enables the model to perform reasoning on the input image. In particular, the captions generated by more CFC layers contain high-level words inferred by some regions and their association with some attributes that encode a semantic concept. For example, the terms "washing" in the lower-left part and "catch" in the upper-right part of Fig. \ref{fig_5} are such high-level words that our model can generate accurately. Furthermore, the phrase "working in a computer lab" in the upper-right part of the figure is an excellent example of generating a conceptual expression through multiple reasoning steps.

It is worth noting that captions with one CFC layer contain both visual and non-visual words regarding contextual information, indicating the strength of a standalone CFC layer. For instance, in the second example of the first row in Fig. \ref{fig_5}, the words "standing" and "looking" as non-visual words, and the words "sink" and "window" as visual words show that our model can selectively attend to both visual and non-visual words resulting better captions.

\subsubsection{Model Ensemble}
Ensembles [52] have been a straightforward and effective way to enhance machine learning systems’ performance for a long time. In deep architectures, one only needs to train multiple models independently on the same task, potentially varying some training circumstances and aggregating their predictions to make the outcome at inference time. For a fair comparison, we also report the ensemble performance of three SCFC models with different stacked SCFC layers in the range of \(1\) to \(3\). The result is presented in Table \ref{tab:table1} as SCFC[\(\Sigma\)]. Compared with each individual model’s results, the ensemble model boosts the performance significantly, which means various semantic-level words need a different number of reasoning steps. This achievement hints at further research on the dynamic number of stacked SCFC layers.

\subsubsection{Parameter Size Analysis}
Increasing the number of stacked SCFC layers can imitate the multi-step reasoning procedure leading to the generation of fine-grained captions. Besides, deepening the SCFC increases the overall parameter size, which can slow down the speed of convergence. To further verify the influence of growing SCFC, we conduct two experiments as follows:

\begin{itemize}
\item{BU-SCFC: Replace our visual detector with the pre-trained bottom-up attention.}
\item{Gwf-SCFC: Replace our word embedding with the pre-trained Glove without fine-tuning.}
\end{itemize}

\begin{figure}[!t]
\centering
\includegraphics[width=0.6\columnwidth,scale=0.5]{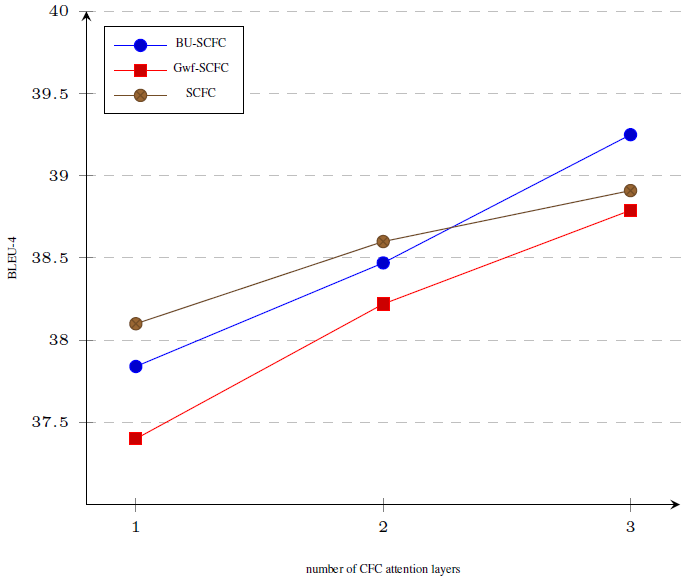}
\caption{The results of ablated models (BU-SCFC[s], Gwf-SCFC[s]) and our entire model SCFC[s] where s denotes the different number of CFC layers on the MSCOCO test split in CIDEr-D score optimization training in terms of BLEU-4.}
\label{fig_6}
\end{figure}

The experimental results are shown in Fig. \ref{fig_6}. The figure shows that increasing the number of SCFCs in the ablative models can further enhance the captioning performance. These improvements significantly occur in the model with three SCFC layers. In particular, the ablative model BU-SCFC achieves a relative improvement of 0.34\% in terms of BLEU-4 compared with the SCFC model with the same stacked SCFC numbers. Although this improvement is remarkable, it comes at the cost of a lack of end-to-end training capability. It should be noted that we do not use these parameter reduction tricks in any of the other comparison results.

\subsubsection{Influence of beam size k}
\begin{figure}[!t]
\centering
\includegraphics[width=\columnwidth]{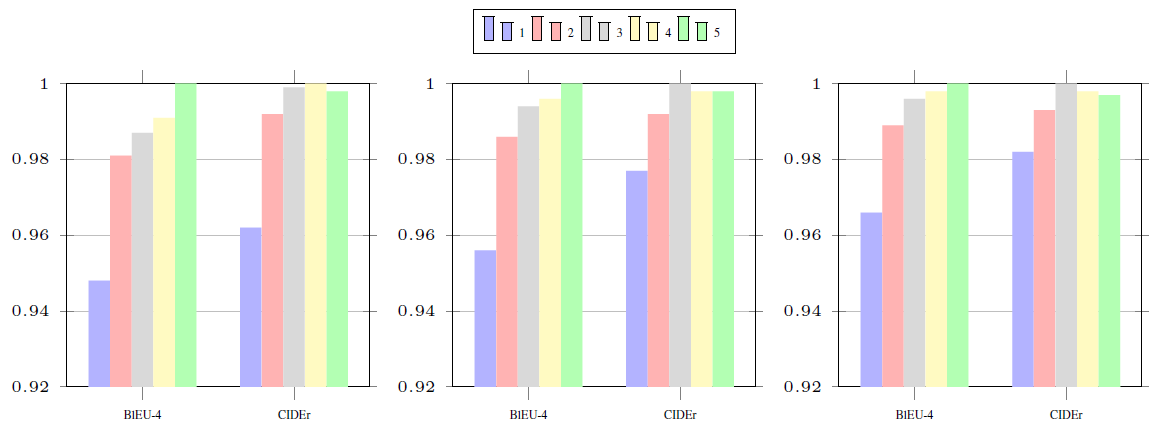}
\caption{The effect of beam size k on BLEU and CIDEr in our proposed SCFC with one/two/three CFC layers from left to right.}
\label{fig_7}
\end{figure}

There are two standard procedures in the test phase to predict the next token in sentence generation. The first one is the greedy method, in which the token with the highest score at each time step is determined and used to predict the next token until it grasps the end flag or the maximum length of the caption. The second one is to use beam search to choose the best \(k\) subsequences at each time step and use them as applicants to generate the best \(k\) subsequences in the next time step. To examine the influence of the beam size in the testing step, we analyze the performances of our grown model with the various numbers of CFC layers with different beam sizes in the common range of \(1\) to \(5\) on the MSCOCO dataset. Fig. \ref{fig_7} shows the results obtained by the normalization step according to the highest score of each evaluation metric. From the figure, we can find that as the beam size \(k\) increases, BLUE-4 will be enhanced in all model variations meaning we can generate sentences fluently by increasing the beam size in the range set in our experiment. In contrast, when the beam size \(k\) is \(3\), the CIDEr metric will reach its peak, except in the model with one CFC layer where the result with the beam size of \(4\) is slightly ahead of the result with the beam size of \(3\).

\section{Conclusion}
This paper presents an SCFC attention network for image captioning to make reasoning in multiple steps on the cross-modal features, thereby achieving a compact and informative representation in the caption generation process. We demonstrate that compounding multimodal information can boost the generation of discriminative features to generate all-level semantic words. We first form the textual component via context-aware attributes, which can jointly train through the captioning network training with the visual detector. We then propose an SCFC attention network to consolidate cross-modal features multiple times, imitating a multi-step reasoning procedure. Besides, we suggest the SCFC-LSTM encourage the model to look at the consolidated representation in a more controlled way. Our SCFC can generate more fine-grained captions. To validate the effectiveness of the proposed method, we conduct extensive experiments. Experimental results show that our method outperforms state-of-the-art models trained by both RL-based and cross-entropy losses. We also plan to design a dynamic SCFC attention network for image captioning tasks when the number of steps parameter depends on the word that should be generated rather than a predefined fixed number.


\section*{References}

\noindent [1] A. Karpathy and L. Fei-Fei, “Deep visual-semantic alignments for generating image descriptions,” in Proceedings of the IEEE conference on computer vision and pattern recognition, 2015, pp. 3128–3137.

\noindent [2] A. Karpathy, A. Joulin, and L. F. Fei-Fei, “Deep fragment embeddings for bidirectional image sentence mapping,” in Advances in neural information processing systems, 2014, pp. 1889–1897.

\noindent [3] H. Liu, Y. Yang, F. Shen, L. Duan, and H. T. Shen, “Recurrent image captioner: Describing images with spatial-invariant transformation and attention filtering,” arXiv preprint arXiv:1612.04949, 2016.

\noindent [4] J. Mao, W. Xu, Y. Yang, J. Wang, Z. Huang, and A. Yuille, “Deep captioning with multimodal recurrent neural networks (m-rnn),” arXiv preprint arXiv:1412.6632, 2015.

\noindent [5] R. Kiros, R. Salakhutdinov, and R. S. Zemel, “Unifying visual-semantic embeddings with multimodal neural language models,” arXiv preprint arXiv:1411.2539, 2014.

\noindent [6] T. J. Buschman and E. K. Miller, “Top-down versus bottom-up control of attention in the prefrontal and posterior parietal cortices,” science, vol. 315, no. 5820, pp. 1860–1862, 2007.

\noindent [7] M. Corbetta and G. L. Shulman, “Control of goal-directed and stimulusdriven attention in the brain,” Nature reviews neuroscience, vol. 3, no. 3, pp. 201–215, 2002.

\noindent [8] K. Xu, J. Ba, R. Kiros, K. Cho, A. Courville, R. Salakhudinov, R. Zemel, and Y. Bengio, “Show, attend and tell: Neural image caption generation with visual attention,” in International conference on machine learning, 2015, pp. 2048–2057.

\noindent [9] Q. You, H. Jin, Z. Wang, C. Fang, and J. Luo, “Image captioning with semantic attention,” in Proceedings of the IEEE conference on computer vision and pattern recognition, 2016, pp. 4651–4659.

\noindent [10] Q. Wu, C. Shen, L. Liu, A. Dick, and A. Van Den Hengel, “What value do explicit high level concepts have in vision to language problems?” in Proceedings of the IEEE conference on computer vision and pattern recognition, 2016, pp. 203–212.

\noindent [11] T. Yao, Y. Pan, Y. Li, Z. Qiu, and T. Mei, “Boosting image captioning with attributes,” in Proceedings of the IEEE International Conference on Computer Vision, 2017, pp. 4894–4902.

\noindent [12] C. He and H. Hu, “Image captioning with text-based visual attention,” Neural Processing Letters, vol. 49, no. 1, pp. 177–185, 2019.

\noindent [13] Y. Su, Y. Li, N. Xu, and A.-A. Liu, “Hierarchical deep neural network for image captioning,” Neural Processing Letters, pp. 1-11, 2019.

\noindent [14] O. Vinyals, A. Toshev, S. Bengio, and D. Erhan, “Show and tell: A neural image caption generator,” in Proceedings of the IEEE conference on computer vision and pattern recognition, 2015, pp. 3156–3164.

\noindent [15] J. Mao, W. Xu, Y. Yang, J. Wang, and A. L. Yuille, “Explain images with multimodal recurrent neural networks,” arXiv preprint arXiv:1410.1090, 2014.

\noindent [16] D. Bahdanau, K. Cho, and Y. Bengio, “Neural machine translation by jointly learning to align and translate,” arXiv preprint arXiv:1409.0473, 2014.

\noindent [17] K. Fu, J. Jin, R. Cui, F. Sha, and C. Zhang, “Aligning where to see and what to tell: Image captioning with region-based attention and scene-specific contexts,” IEEE transactions on pattern analysis and machine intelligence, vol. 39, no. 12, pp. 2321–2334, 2017.

\noindent [18] J. R. Uijlings, K. E. Van De Sande, T. Gevers, and A. W. Smeulders, “Selective search for object recognition,” International journal of computer vision, vol. 104, no. 2, pp. 154–171, 2013.

\noindent [19] M. Pedersoli, T. Lucas, C. Schmid, and J. Verbeek, “Areas of attention for image captioning,” in Proceedings of the IEEE international conference on computer vision, 2017, pp. 1242–1250.

\noindent [20] J. Lu, C. Xiong, D. Parikh, and R. Socher, “Knowing when to look: Adaptive attention via a visual sentinel for image captioning,” in Proceedings of the IEEE conference on computer vision and pattern recognition, 2017, pp. 375–383.

\noindent [21] Y. Huang, J. Chen, W. Ouyang, W. Wan, and Y. Xue, “Image captioning with end-to-end attribute detection and subsequent attributes prediction,” IEEE Transactions on Image Processing, vol. 29, pp. 4013–4026, 2020.

\noindent [22] P. Anderson, X. He, C. Buehler, D. Teney, M. Johnson, S. Gould, and L. Zhang, “Bottom-up and top-down attention for image captioning and visual question answering,” in Proceedings of the IEEE conference on computer vision and pattern recognition, 2018, pp. 6077–6086.

\noindent [23] K. He, X. Zhang, S. Ren, and J. Sun, “Deep residual learning for image recognition,” in Proceedings of the IEEE conference on computer vision and pattern recognition, 2016, pp. 770–778.

\noindent [24] S. Ren, K. He, R. Girshick, and J. Sun, “Faster r-cnn: Towards real-time object detection with region proposal networks,” in Advances in neural information processing systems, 2015, pp. 91–99.

\noindent [25] J. Johnson, A. Karpathy, and L. Fei-Fei, “Densecap: Fully convolutional localization networks for dense captioning,” in Proceedings of the IEEE Conference on Computer Vision and Pattern Recognition, 2016.

\noindent [26] M. Jaderberg, K. Simonyan, A. Zisserman et al., “Spatial transformer networks,” in Advances in neural information processing systems, 2015, pp. 2017–2025.

\noindent [27] H. Fang, S. Gupta, F. Iandola, R. K. Srivastava, L. Deng, P. Dollár, J. Gao, X. He, M. Mitchell, J. C. Platt et al., “From captions to visual concepts and back,” in Proceedings of the IEEE conference on computer vision and pattern recognition, 2015, pp. 1473–1482.

\noindent [28] T.-Y. Lin, P. Goyal, R. Girshick, K. He, and P. Dollár, “Focal loss for dense object detection,” in Proceedings of the IEEE international conference on computer vision, 2017, pp. 2980–2988.

\noindent [29] F. A. Gers and J. Schmidhuber, “Recurrent nets that time and count,” in Proceedings of the IEEE-INNS-ENNS International Joint Conference on Neural Networks. IJCNN 2000. Neural Computing: New Challenges and Perspectives for the New Millennium, vol. 3. IEEE, 2000, pp. 189–194.

\noindent [30] K. Greff, R. K. Srivastava, J. Koutník, B. R. Steunebrink, and J. Schmidhuber, “Lstm: A search space odyssey,” IEEE transactions on neural networks and learning systems, vol. 28, no. 10, pp. 2222–2232, 2016.

\noindent [31] R. Vedantam, C. Lawrence Zitnick, and D. Parikh, “Cider: Consensus-based image description evaluation,” in Proceedings of the IEEE conference on computer vision and pattern recognition, 2015, pp. 4566–4575.

\noindent [32] S. J. Rennie, E. Marcheret, Y. Mroueh, J. Ross, and V. Goel, “Selfcritical sequence training for image captioning,” in Proceedings of the IEEE Conference on Computer Vision and Pattern Recognition, 2017, pp. 7008–7024.

\noindent [33] T.-Y. Lin, M. Maire, S. Belongie, J. Hays, P. Perona, D. Ramanan, P. Dollár, and C. L. Zitnick, “Microsoft coco: Common objects in context,” in European conference on computer vision. Springer, 2014, pp. 740–755.

\noindent [34] P. Young, A. Lai, M. Hodosh, and J. Hockenmaier, “From image descriptions to visual denotations: New similarity metrics for semantic inference over event descriptions,” Transactions of the Association for Computational Linguistics, vol. 2, pp. 67–78, 2014.

\noindent [35] K. Papineni, S. Roukos, T. Ward, and W.-J. Zhu, “Bleu: a method for automatic evaluation of machine translation,” in Proceedings of the 40th annual meeting of the Association for Computational Linguistics, 2002, pp. 311–318.

\noindent [36] M. Denkowski and A. Lavie, “Meteor universal: Language-specific translation evaluation for any target language,” in Proceedings of the ninth workshop on statistical machine translation, 2014, pp. 376–380.

\noindent [37] C.-Y. Lin, “Rouge: A package for automatic evaluation of summaries,” in Text summarization branches out, 2004, pp. 74 81.

\noindent [38] P. Anderson, B. Fernando, M. Johnson, and S. Gould, “Spice: Semantic propositional image caption evaluation,” in European Conference on Computer Vision. Springer, 2016, pp. 382–398.

\noindent [39] Z. Gan, C. Gan, X. He, Y. Pu, K. Tran, J. Gao, L. Carin, and L. Deng, “Semantic compositional networks for visual captioning,” in Proceedings of the IEEE conference on computer vision and pattern recognition, 2017, pp. 5630–5639.

\noindent [40] J. Deng, W. Dong, R. Socher, L.-J. Li, K. Li, and L. Fei-Fei, “Imagenet: A large-scale hierarchical image database,” in 2009 IEEE conference on computer vision and pattern recognition. Ieee, 2009, pp. 248–255.

\noindent [41] D. P. Kingma and J. Ba, “Adam: A method for stochastic optimization,” arXiv preprint arXiv:1412.6980, 2014.

\noindent [42] N. Srivastava, G. Hinton, A. Krizhevsky, I. Sutskever, and R. Salakhutdinov, “Dropout: a simple way to prevent neural networks from overfitting,” The journal of machine learning research, vol. 15, no. 1, pp. 1929–1958, 2014.

\noindent [43] A. Graves, “Generating sequences with recurrent neural networks,” arXiv preprint arXiv:1308.0850, 2013.

\noindent [44] R. J. Williams and D. Zipser, “A learning algorithm for continually running fully recurrent neural networks,” Neural computation, vol. 1, no. 2, pp. 270–280, 1989.

\noindent [45] L. Chen, H. Zhang, J. Xiao, L. Nie, J. Shao, W. Liu, and T.-S. Chua, “Sca-cnn: Spatial and channel-wise attention in convolutional networks for image captioning,” in Proceedings of the IEEE conference on computer vision and pattern recognition, 2017, pp. 5659–5667.

\noindent [46] W. Jiang, L. Ma, Y.-G. Jiang, W. Liu, and T. Zhang, “Recurrent fusion network for image captioning,” in Proceedings of the European Conference on Computer Vision (ECCV), 2018, pp. 499–515.

\noindent [47] T. Yao, Y. Pan, Y. Li, and T. Mei, “Exploring visual relationship for image captioning,” in Proceedings of the European conference on computer vision (ECCV), 2018, pp. 684–699.

\noindent [48] J. Wang, W. Wang, L.Wang, Z.Wang, D. D. Feng, and T. Tan, “Learning visual relationship and context-aware attention for image captioning,” Pattern Recognition, vol. 98, p. 107075, 2020.

\noindent [49] X. Yang, K. Tang, H. Zhang, and J. Cai, “Auto-encoding scene graphs for image captioning,” in Proceedings of the IEEE Conference on Computer Vision and Pattern Recognition, 2019, pp. 10 685–10 694.

\noindent [50] Z.-J. Zha, D. Liu, H. Zhang, Y. Zhang, and F. Wu, “Context aware visual policy network for fine-grained image captioning,” IEEE transactions on pattern analysis and machine intelligence, 2019.

\noindent [51] Y. Qin, J. Du, Y. Zhang, and H. Lu, “Look back and predict forward in image captioning,” in Proceedings of the IEEE Conference on Computer Vision and Pattern Recognition, 2019, pp. 8367–8375.

\noindent [52] L. Breiman, “Bagging predictors,” Machine learning, vol. 24, no. 2, pp. 123–140, 1996.

\end{document}